\documentclass{article}

\usepackage[final]{neurips_2019}

\usepackage[utf8]{inputenc} 
\usepackage[T1]{fontenc}    
\usepackage{hyperref}       
\usepackage{url}            
\usepackage{booktabs}       
\usepackage{amsfonts}       
\usepackage{nicefrac}       
\usepackage{microtype}      
\usepackage{bm}      
\usepackage{graphicx}
\usepackage{subcaption}
\usepackage[title]{appendix}

\title{Explainable Prediction of Adverse Outcomes Using Clinical Notes}

\author{%
   Justin R. Lovelace \\
   Texas A\&M University \\
   \And
   Nathan C. Hurley \\
   Texas A\&M University \\
   \AND
   Adrian D. Haimovich \\
   Yale School of Medicine \\
   \And
   Bobak J. Mortazavi \\
   Texas A\&M University \\
}

\begin{document}

\maketitle

\begin{abstract}
  Clinical notes contain a large amount of clinically valuable information that is ignored in many clinical decision support systems due to the difficulty that comes with mining that information. Recent work has found success leveraging deep learning models for the prediction of clinical outcomes using clinical notes. However, these models fail to provide clinically relevant and interpretable information that clinicians can utilize for informed clinical care. In this work, we augment a popular convolutional model with an attention mechanism and apply it to unstructured clinical notes for the prediction of ICU readmission and mortality. We find that the addition of the attention mechanism leads to competitive performance while allowing for the straightforward interpretation of predictions. We develop clear visualizations to present important spans of text for both individual predictions and high-risk cohorts. We then conduct a qualitative analysis and demonstrate that our model is consistently attending to clinically meaningful portions of the narrative for all of the outcomes that we explore.
\end{abstract}

\section{Introduction}

The intensive care unit (ICU) provides care for hospital patients with serious illnesses or injuries. Intensive care is extremely resource intensive so decisions to discharge patients from the ICU must be made in order to allocate limited resources efficiently. An accurate predictor for the likelihood of adverse outcomes would be useful for planning the ICU discharge process and determining the level of monitoring that the patient would receive after discharge from the ICU. In this work, we focus on predicting the likelihood of ICU readmission and mortality using information available at the time of ICU discharge.

Previous work has found success using the large amount of data stored in electronic health records (EHR) to develop predictive models for ICU readmission and mortality \citep{arash, Ghassemi2014, FIALHO, nomogram, saps, icu_readmission_death}. Much of the work in this area is conducted using the publicly available Medical Information Mart for Intensive Care (MIMIC-III) database which is widely considered the gold standard dataset in critical care, due to the volume and quality of the data it contains \citep{mimic}. This work is also performed upon the MIMIC-III database to allow for reproducibility and to encourage future work.


The contributions of this work are as follows. First, we extend a convolutional model commonly used for text classification by augmenting it with an attention layer and apply it to predict ICU readmission and patient mortality. We demonstrate that our model performs competitively while allowing for clearer interpretability due to the addition of the attention mechanism. Second, we provide intuitive visualizations that reveal global trends for each outcome as well as patient-specific visualizations that reveal what sections of the text were important for individual predictions. Third, we conduct an in-depth qualitative analysis of what the model attends to for high-risk cases for all the outcomes that we explore in this work. In doing so, we verify that our model is successfully attending to clinically informative sections of the narrative.

\section{Related work}

There has been some recent work that looks at predicting readmission and mortality by treating them as a single outcome \citep{analysis_attn_clinical, Lin}. However, clinical work by \citet{harlan} has demonstrated that these are orthogonal outcomes, and thus modeling them jointly as a single outcome does not make sense from a clinical perspective. By treating them as separate outcomes in this work, we are able to independently explore the risk factors for these two distinct outcomes. We focus on convolutional architecture in this work due to the work of \citet{serrano} who found that the interpretation of convolutional models was more straightforward than recurrent ones.

\citet{analysis_attn_clinical} also conducted experiments to explore the interpretability of attention for clinical tasks. Their experiments, however, did nothing to explore the quality of the explanations produced by the model as this would require human analysis (further discussion of their work and the interpretability of attention can be found in the appendix). In our work, however, we demonstrate through an in-depth qualitative analysis that the attention mechanism is effectively providing a clinically informative, meaningful explanation for its predictions.

\section{Data and cohort}

This work is conducted using the free text notes stored in the publicly available MIMIC-III database \citep{mimic}. To develop our cohort, we first filter out minors because children have different root causes for adverse medical outcomes than the general populace. We also remove patients who died while in the ICU and filter out ICU stays that are missing information regarding the time of admission or discharge. We then extract all ICU stays where the patient had at least three notes on record before the time of ICU discharge. This leaves us with $33,311$ unique patients and $45,260$ ICU stays.

For ICU readmission we extract labels for two types of readmissions, bounceback and 30 day readmisssion. Bounceback readmissions occur when a patient is discharged from the ICU and then readmitted to the ICU before being discharged from the hospital. For 30 day readmissions, we simply look at any readmission to the ICU within the 30 days following ICU discharge. For mortality, we also look at two different outcomes, in-hospital mortality and 30-day mortality. This provides us with a cohort with $3,413$ bounceback readmissions, $5,674$ 30-day readmissions, $3,761$ deaths within 30 days, and $1,898$ in-hospital deaths. 

\section{Methods}
This work explored two different methods for developing word embeddings. The first method explored is the Word2Vec algorithm using the continuous bag-of-words implementation introduced by \citet{w2v}. We also explore the StarSpace embedding algorithm introduced by \citet{wu2017starspace} which allows for the training of word embeddings based on document labels. We follow the training setup of \citet{nyu} and treat each patient's clinical narrative as a bag-of-words and then label it with the ICD codes assigned to their hospital stay to learn the word embeddings.

The CNN architecture used in this work was originally developed by \citet{CNN} for the task of sentence classification and has since been used with clinical notes for a variety of different tasks \citep{nyu, residual, multimodal}. For our baseline model we follow the work of \citet{CNN} and apply a max-pooling operation to the feature maps produced by applying each filter to the entire clinical narrative. In this work, we use three convolutional filters of width 1, 2, and 3. The three max pooled representations for the filters are then concatenated together to produce the final feature representation that is used for the prediction.

In this work, we also augment the CNN with a scaled dot-product attention mechanism \citep{attn_all_you_need}. Unlike the work of \cite{caml}, we apply our attention mechanism over multiple convolutional filters of different lengths, rather than a single filter of static length. This allows our model to consider variable spans of text whereas their model was limited to a fixed context window. To do so, we learn a query vector, $\mathbf{q} \in \mathbb{R}^{j}$, that will be used to calculate the importance of the feature maps across all filters using the dot product operation. We concatenate the output of all of the convolutional filters into a single matrix $\mathbf{H}$. We can then calculate the attention distribution using the matrix vector product of our final feature map and the query vector, $\boldsymbol{\alpha}=softmax(\frac{\mathbf{H^{T}}\mathbf{q}}{\sqrt{d}})$, where $d$ is the output dimensionality of the filters and $\boldsymbol{\alpha}$ contains the score for every position across all the filters. We calculate the final representation used for classification by taking a weighted average of all of the outputs based on their calculated weights $\mathbf{v} = \sum_{i=1}^{3N}\boldsymbol{\alpha_i}\mathbf{h_i}$ where $\mathbf{v}$ is the final representation used for the prediction.

Given the final patient representation, we calculate the final prediction using a sigmoid layer that computes the likelihood for the targeted outcome. We train our classifier by minimizing the binary cross-entropy loss function.  Further implementation details can be found in the appendix.

\section{Experiments and results}
We conduct all of our experiments using both the max pooled CNN (CNN-Max) and the CNN augmented with an attention mechanism (CNN-Attn). We train each model using both the Word2Vec and the StarSpace embeddings and evaluate them on the two readmission and the two mortality tasks. For all of our experiments we report both the Area Under the Receiver Operator Curve (AU-ROC) and the Area Under the Precision Recall Curve (AU-PR). All of the metrics reported in this work are the average performance over 5-fold cross validation using an 80/10/10 split based on the subject ID. We report the standard deviation over the folds as confidence bounds.

The results for the readmission outcomes are reported in table \ref{readmission-table} and the results for the mortality outcomes are reported in table \ref{mortality-table}. Our findings demonstrates that the addition of the attention mechanism leads to comparable, and in some cases improved, performance when compared against a strong baseline that has been used in recent work in the clinical domain. We also demonstrate that while StarSpace embeddings do improve the performance of the CNN-Max model across all tasks, this trend does not hold when altering the architecture. Because of this, it appears the optimal choice of embedding technique may be architecture dependent.
\begin{table}
  \caption{ Readmission Results}
  \label{readmission-table}
  \centering
  \begin{tabular}{llllll}
    \toprule
    \multicolumn{2}{c}{Method} & \multicolumn{2}{c}{30-Day Readmission} & \multicolumn{2}{c}{Bounceback Readmission}       \\
    \cmidrule(r){1-6} 
     Model     & Embedding & AU-ROC     & AU-PR & AU-ROC     & AU-PR\\ 
    
    \midrule
    CNN-Max & Word2Vec  &    $0.65 \pm 0.01$   & $0.21 \pm 0.02$ & $0.66 \pm 0.02$  & $0.15 \pm 0.02$\\
    CNN-Attn     & Word2Vec &   $\mathbf{0.68 \pm 0.00}$    & $\mathbf{0.24 \pm 0.02}$ &  $\mathbf{0.71 \pm 0.01}$ & $\mathbf{0.17 \pm 0.02}$  \\
    CNN-Max     & StarSpace       &    $\mathbf{0.68 \pm 0.00}$   & $0.23 \pm 0.01$ & $0.69 \pm 0.01$  & $\mathbf{0.17 \pm 0.02}$\\
    CNN-Attn     & StarSpace       &  $\mathbf{0.68 \pm 0.00}$   & $0.23 \pm 0.02$ &  $0.70 \pm 0.01$ & $0.16 \pm 0.02$ \\
    \bottomrule
  \end{tabular}
\end{table}

\begin{table}
  \caption{ Mortality Results}
  \label{mortality-table}
  \centering
  \begin{tabular}{llllll}
    \toprule
    \multicolumn{2}{c}{Method} & \multicolumn{2}{c}{30-Day Mortality} & \multicolumn{2}{c}{In-Hospital Mortality}       \\
    \cmidrule(r){1-6} 
     Model     & Embedding & AU-ROC     & AU-PR & AU-ROC     & AU-PR\\ 
    
    \midrule
    CNN-Max & Word2Vec  & $0.84 \pm 0.01$ & $0.43 \pm 0.01$ &  $0.85 \pm 0.01$ &  $0.32 \pm 0.05$ \\
    CNN-Attn     & Word2Vec & $0.85 \pm 0.01$  & $0.42 \pm 0.01$ &  $\mathbf{0.87 \pm 0.01}$ &  $0.33 \pm 0.04$  \\
    CNN-Max     & StarSpace & $\mathbf{0.86 \pm 0.01}$ & $\mathbf{0.45 \pm 0.01}$ & $\mathbf{0.87 \pm 0.01}$ & $\mathbf{0.34 \pm 0.05}$ \\
    CNN-Attn     & StarSpace & $\mathbf{0.86 \pm 0.01}$ & $0.43 \pm 0.01$ & $0.86 \pm 0.01$ & $0.30 \pm 0.05$  \\
    \bottomrule
  \end{tabular}
\end{table}

\section{Interpretability} \label{interpretability}

\begin{figure}
  \begin{subfigure}[t]{0.5\linewidth}
    \centering
    \fbox{\includegraphics[width=.8\linewidth]{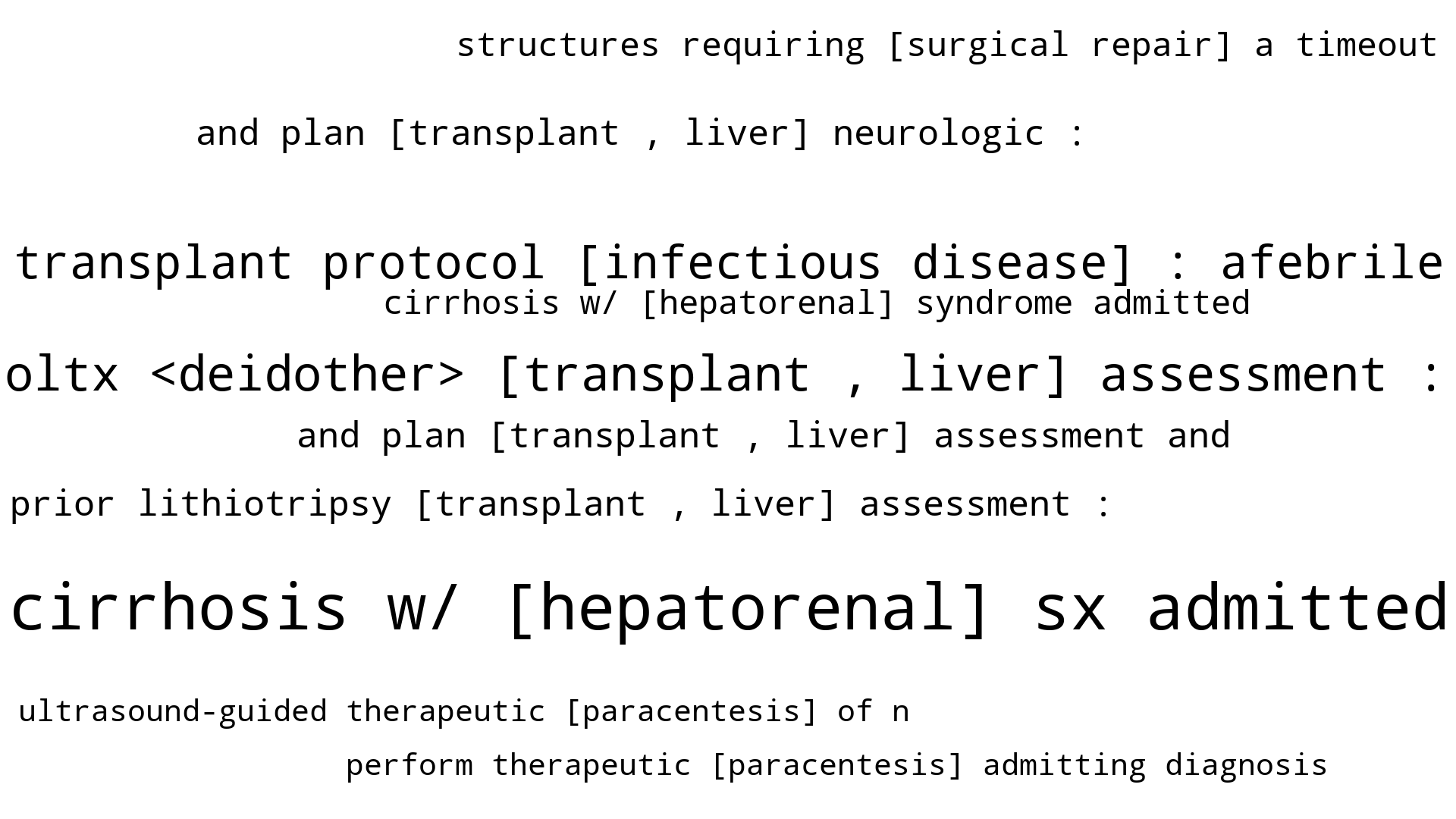}}
    \caption{Patient at high risk of bounceback readmission.}
    \label{fig:individual-wordcloud}
  \end{subfigure}
  \begin{subfigure}[t]{0.5\linewidth}
    \centering
    \fbox{\includegraphics[width=.8\linewidth]{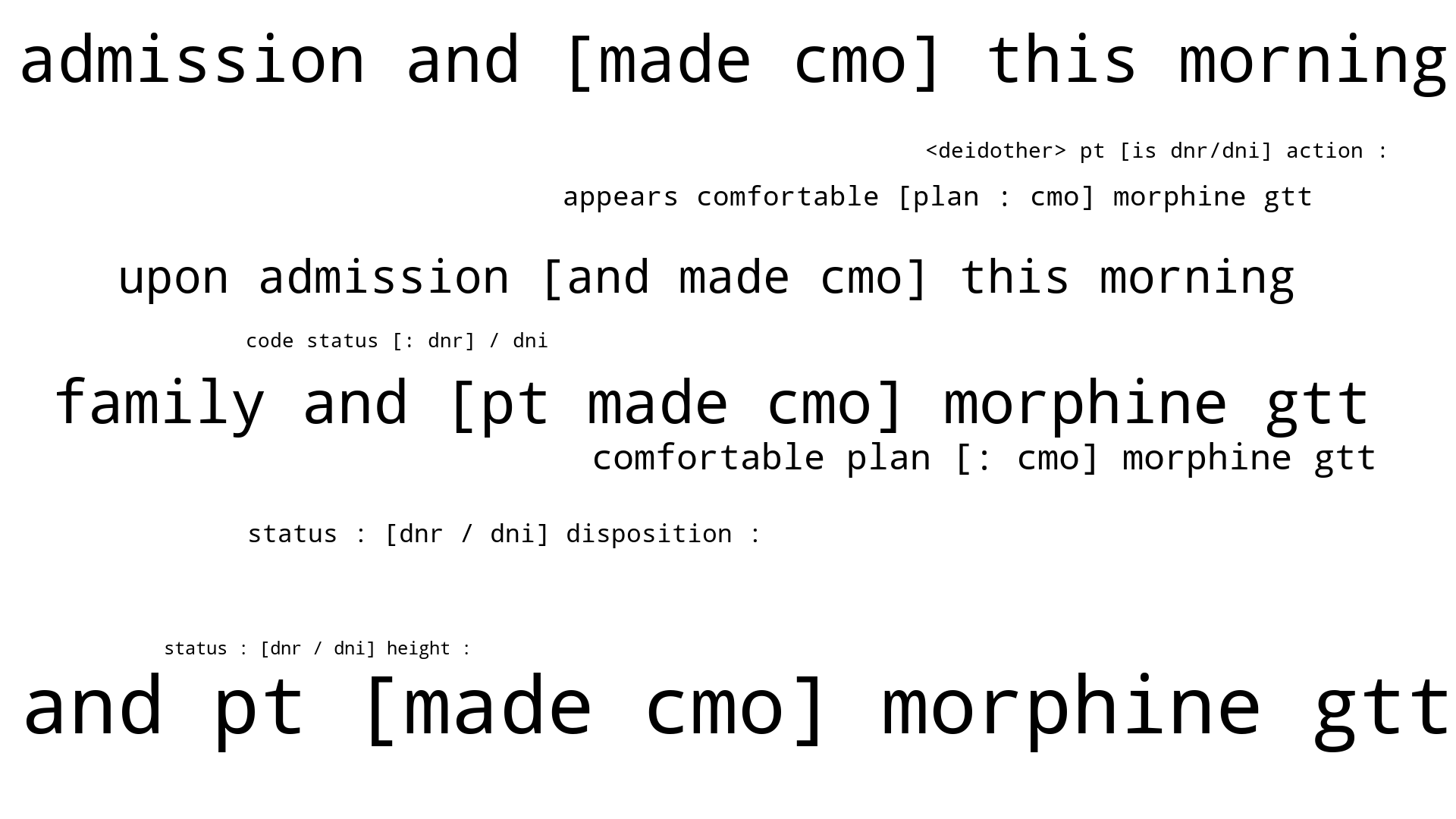}}
    \caption{Patient at high risk of in-hospital mortality.}
    \label{fig:global-wordcloud}
  \end{subfigure}
\caption{Visualizations of attention distributions for individual predictions}
\label{fig:individual_viz}
\end{figure}

\begin{figure}
  \begin{subfigure}[t]{0.5\linewidth}
    \centering
    \fbox{\includegraphics[width=.8\linewidth]{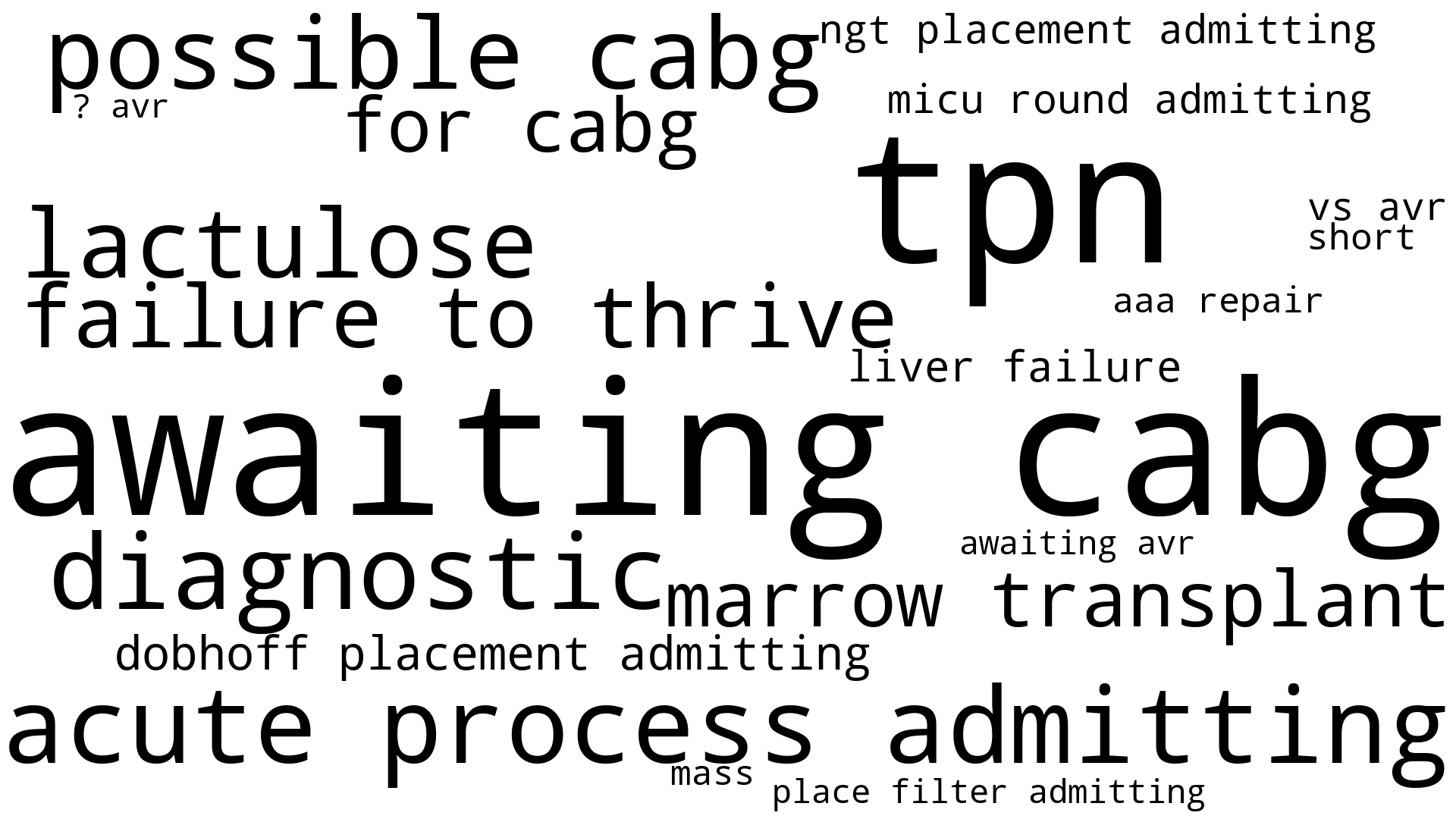}}
    \caption{High-risk cohort for bounceback readmission.}
    \label{fig:global-wordcloud}
  \end{subfigure}
  \begin{subfigure}[t]{0.5\linewidth}
    \centering
    \fbox{\includegraphics[width=.8\linewidth]{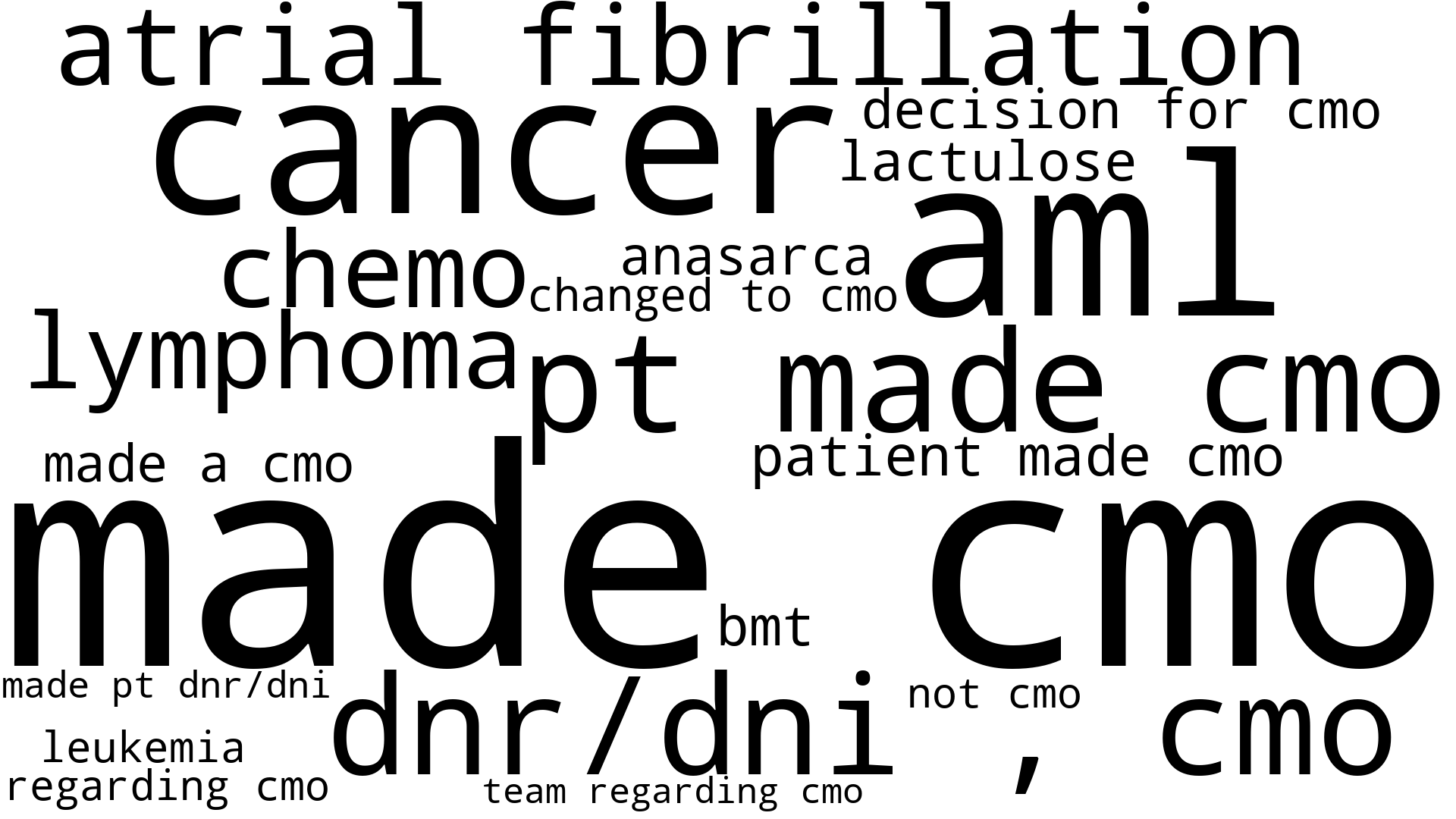}}
    \caption{High-risk cohort for in-hospital mortality.}
    \label{fig:ind_mort}
  \end{subfigure}
\caption{Visualizations of attention distributions for high-risk cohorts}
\label{fig:global_viz}
\end{figure}

We construct visualizations to highlight what portions of the narrative are being attended to for both individual predictions (figure \ref{fig:individual_viz}) and high-risk cohorts (figure \ref{fig:global_viz}). Details regarding how these visualizations were constructed and discussion on their interpretation can be found in the appendix. We find that our visualizations reveal clear, clinically informative trends for each outcome.

Likelihood of readmission is found to be related to words describing conditions such as liver and heart disease. There are many refences to end-stage liver disease, as well as to words suggesting multiple comorbidities and complications.  For instance, we find multiple phrases similar to ``multiple admissions'' or ``complicated by''.  Notably, references to malignancies are largely absent.  In addition to sharing risk factors with 30-day readmission, high-risk of bounceback readmission is found near words suggestive of more involved care.  For instance, mentions of interventions such as central lines or infectious disease are associated with bounceback readmission.

For both mortality outcomes, the predictions for the highest risk patients are unsurprisingly driven largely by references to end-of-life care. Of more clinical interest, we observe many references to cancer and descriptors that are associated with poor cancer outcomes (such as metastasis). Additionally, we also see words correlated with severe liver disease and atrial fibrillation, which is associated with negative outcomes such as stroke and heart failure.


\section{Conclusion}
In this work, we augment a popular convolutional model with an attention mechanism and demonstrate that it achieves competitive performance with a strong baseline that has been used in recent work in this domain. Additionally, we conduct a qualitative analysis and find that the attention mechanism is able to consistently identify clinically pertinent information contained within the clinical narrative. The ability to provide insight into what led our model to make its decision is particularly valuable in the clinical domain where transparency is important for building trust with clinicians. We provide easy ways to visualize important spans of text for both individual predictions and high-risk cohorts and utilize visualizations of the latter to explore what risk factors are associated with the outcomes explored in this work. 

In the future, we intend to extend this work to a variety of different prediction time frames and explore whether we can identify distinct risk factors between short and long term readmission and mortality outcomes. We also intend to explore making the prediction from an earlier point in the hospital stay to improve the clinical utility of our model.
\medskip

\small
\bibliography{references}
\bibliographystyle{plainnat}

\begin{appendices}
\section{Interpretability of attention}
\citet{analysis_attn_clinical} recently explored the effect of augmenting LSTMs with an additive attention mechanism for the prediction of various clinical outcomes using notes. They also performed two experiments to explore the ability of the attention mechanism to explain their model predictions. In the first experiment they found that the attention distributions correlated weakly, but almost always statistically significantly, with gradient measures of feature importance. Although prior work by \citet{gradient}. demonstrates that there are also problems with using gradient measures for interpretation, we believe their work does raise potential problems with the interpretability of attentive recurrent encoders. They, however, also demonstrate that attention distributions over a simpler embedding encoder baseline consistently had a much stronger correlation with the gradient measure of importance. Because of this second finding, we use convolutional architecture, which is closely related to the embedding encoder, rather than recurrent architecture in this work. This choice is also supported by the work of \citet{serrano}.

\citeauthor{analysis_attn_clinical} also conducted a second experiment that they had introduced in prior work and claimed that it demonstrates that attention mechanisms could not reliably be treated as explanatory \citep{attn_not}. However, \citet{attn_not_not} explored some of the problems with their underlying assumptions and experimental setup and demonstrated that their experiment failed to fully explore their premise, and thus failed to support their claim. While complete details can be found by referencing those works, we also take some issue with the premise that motivates their experiment. They claim that for attention to be explanatory, it must be exclusive.

However, we agree with the claim made by \citeauthor{attn_not_not} that explainability does not entail exclusivity.  Even if the final output of the model could be aggregated differently to come to a similar conclusion, the model is still making a choice using the trained attention mechanism to weigh the output in the way that it does. The existence of a possible alternative does not change the fact that the model was trained to weigh the text in a certain way, and this can provide insight into what the model views as important. Their exploration of attention also did nothing to evaluate the quality of the explanations through the lens of human analysis, something that is essential for verifying the ability of a model to identify clinically meaningful trends. We address this deficiency in our work by conducting a qualitative analysis and demonstrating that our models are consistently attending to clinically meaningful portions of the narrative.

\section{Implementation details}

\subsection{Data processing}
We extract all clinical notes associated with a patient's hospital stay up until the time of their discharge from the ICU. The text is then preprocessed by lowercasing the text, normalizing punctuation, and replacing numerical characters and de-identified information with generic tokens. All of the notes for each patient are then concatenated and treated as a continuous sequence of text which is used as the input to all of our models. We truncate or pad all clinical narratives to 8000 tokens. This captures the entire clinical narrative for over $75\%$ of patients and we found that extending the maximum sequence length beyond that point did not lead to any further improvements in performance.

\subsection{Word embeddings}
We utilize all notes in the MIMIC-III database associated with subjects who are not in our testing set to train embeddings. This allows for training on a greater selection of notes than if training had been limited to the training set. We generate embeddings for all words that appear in at least 5 documents in our training corpus. Both 100 and 300 dimensional word embeddings were explored and we used 100 dimensions for Word2Vec embeddings and 300 dimensions for StarSpace embeddings based on early testing. We replace out-of-vocabulary words with a randomly initialized UNK token to represent unknown words. This is used to represent words that are not seen in the training data or that are too uncommon for the embedding techniques to effectively learn meaningful representations.

\subsection{Hyperparameters}
The output dimensionality of all of our convolutional filters is set to $64$. We apply dropout with a probability of $0.2$ after the embedding layer and apply it with a probability of $0.3$ after the convolutional layer and after the max pooling or attention layer. All of our models were trained with a batch size of 32 using a learning rate of $0.001$ using the Adam optimizer \citep{adam}. We stop training when the Area Under the Receiver Operator Curve has not improved for 10 epochs and then evaluate the model with the best validation performance on the test set. All of our hyperparameters were selected based on validation performance. Our code is made publicly available at \url{https://github.com/justinlovelace/explainable-mimic-predictions}.

\section{Visualizations}

For the visualization of individual predictions, we construct a word cloud of the $10$ most important phrases whose sizes are scaled based on their attention weights. The words that are attended to are bracketed and neighboring words are provided for further context. An example of a patient at a high risk for bounceback readmission and a patient at a high risk for in-hospital mortality can be seen in figure \ref{fig:individual_viz}. The figure demonstrates that the model predicts the first patient to be at a high risk of readmission because of severe liver disease and its associated comorbidities. The second patient is at a high risk of mortality because he was made cmo (or "comfort measures only") which refers to the treatment plan where doctors allow the patient to die naturally while only ensuring comfort. While this second prediction is an obvious case, it does demonstrate that the model is effectively focusing on the most important parts of the narrative for the predictive task.

Additionally, we explore visualizing high-risk cohorts as a group to elucidate global trends. Similar to the individual case, we do so by constructing a wordcloud; we aggregate the attention weights by summing the weights for every occurrence of a phrase that is in one of the top $3$ phrases for patients in the cohort. The wordcloud for the top $5\%$ of in-hospital mortality and bounceback readmission predictions for one of the testing folds can be seen in figure \ref{fig:global_viz}. The trends found in high-risk cohorts for the readmission and mortality outcomes are discussed in section \ref{interpretability}.

\end{appendices}

\end{document}